\documentclass[10pt,twocolumn,letterpaper]{article}

\usepackage[pagenumbers]{cvpr} 

\usepackage{graphicx}
\usepackage{amsmath}
\usepackage{mathtools}
\usepackage{amssymb}
\usepackage{booktabs}
\usepackage{csquotes}
\usepackage[dvipsnames]{xcolor}
\DeclareMathAlphabet\mathbfcal{OMS}{cmsy}{b}{n}
\usepackage{multirow}
\usepackage{diagbox}
\usepackage{enumitem}

\usepackage[pagebackref,breaklinks,colorlinks]{hyperref}

\usepackage[capitalize]{cleveref}
\crefname{section}{Sec.}{Secs.}
\Crefname{section}{Section}{Sections}
\Crefname{table}{Table}{Tables}
\crefname{table}{Tab.}{Tabs.}

\newcommand{\bm}{\mathbf{m}}

\newcommand{\bp}{\mathbf{p}}

\newcommand{\cS}{\mathcal{S}}

\makeatletter
\DeclareRobustCommand\onedot{\futurelet\@let@token\@onedot}
\def\@onedot{\ifx\@let@token.\else.\null\fi\xspace}
\def\eg{e.g\onedot} 
\def\ie{i.e\onedot}

\def\etal{et~al\onedot}

\makeatother

\definecolor{darkgreen}{rgb}{0,0.7,0}

\def\cvprPaperID{8335} 
\def\confName{CVPR}
\def\confYear{2023}

\begin{document}

\title{\vspace{-0.2cm} Attention-Based Point Cloud Edge Sampling}

\author{Chengzhi Wu$^{1}$ \quad Junwei Zheng$^{1}$
\quad Julius Pfrommer$^{2,3}$ \quad Jürgen Beyerer$^{1,2,3}$ 
\and
$^{1}$Karlsruhe Institute of Technology, Germany \quad
$^{2}$Fraunhofer IOSB, Germany\\
$^{3}$Fraunhofer Center for Machine Learning, Germany \vspace{-0.2cm}
\and
{\tt\footnotesize
\{chengzhi.wu, junwei.zheng\}@kit.edu, \quad \{julius.pfrommer, juergen.beyerer\}@iosb.fraunhofer.de}
\vspace{-0.1cm}}

\maketitle

\begin{abstract}
Point cloud sampling is a less explored research topic for this data representation. The most commonly used sampling methods are still classical random sampling and farthest point sampling. With the development of neural networks, various methods have been proposed to sample point clouds in a task-based learning manner. However, these methods are mostly generative-based, rather than selecting points directly using mathematical statistics. Inspired by the Canny edge detection algorithm for images and with the help of the attention mechanism, this paper proposes a non-generative Attention-based Point cloud Edge Sampling method (APES), which captures salient points in the point cloud outline. Both qualitative and quantitative experimental results show the superior performance of our sampling method on common benchmark tasks. 
\vspace{-0.2cm}
\end{abstract}

\section{Introduction}
\label{sec:intro}
Point clouds are a widely used data representation in various domains including autonomous driving, augmented reality, and robotics. Due to the typically large amount of data, the sampling of a representative subset of points is a fundamental and important task in 3D computer vision.

Apart from random sampling (RS), other classical point sampling methods including grid sampling, uniform sampling, and geometric sampling have been well-established. 
Grid sampling samples points with regular grids and thus cannot control the number of sampled points exactly.
Uniform sampling takes the points in the point cloud evenly and is more popular due to its robustness. Farthest point sampling (FPS) \cite{Eldar1994TheFP, Moenning2003FastMF} is the most famous of them and has been widely used in many current methods when downsampling operations are required \cite{Qi2017PointNetDH, Yu2018PUNetPC, Li2018PointCNNCO, Wu2019PointConvDC, Yan2020PointASNLRP}.
Geometric sampling samples points based on local geometry, such as the curvature of the underlying shape.
Another example of Inverse Density Importance Sampling (IDIS)\cite{Groh2018FlexConvolutionM} samples points whose distance sum values with neighbors are smaller. But this method requires the point cloud to have a high density throughout, and it performs even worse when the raw point cloud has an uneven distribution. 

\begin{figure}[t]
    \centering
    \includegraphics[width=1\linewidth,trim=2 10 2 10,clip]{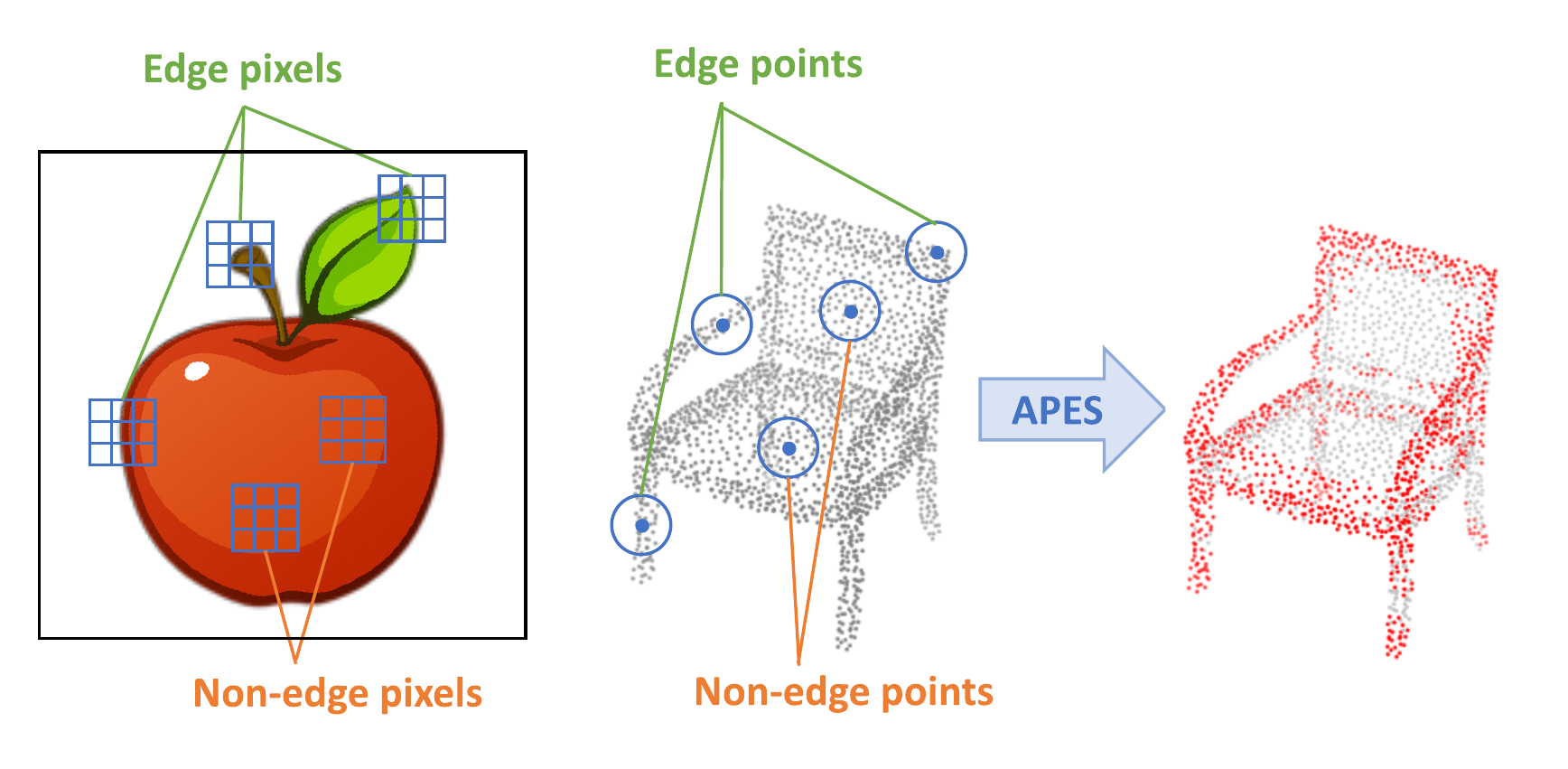}
    \caption{Similar to the Canny edge detection algorithm that detects edge pixels in images, our proposed APES algorithm samples edge points which indicate the outline of the input point clouds. The blue grids/spheres represent the local patches for given center pixels/points.}
    \label{fig:canny}
\end{figure}

In addition to the above mathematical statistics-based methods, with the development of deep learning techniques, several neural network-based methods have been proposed for task-oriented sampling, including S-Net \cite{Dovrat2019LearningTS}, SampleNet \cite{Lang2020SampleNetDP}, DA-Net \cite{Lin2021DANetDD}, etc. They use simple multi-layer perceptrons (MLPs) to generate new point cloud sets of desired sizes as resampled results, supplemented by different post-processing operations. MOPS-Net \cite{Qian2020MOPSNetAM} learns a sampling transformation matrix first, and then generates the sampled point cloud by multiplying it with the original point cloud. However, all these methods are generative-based, rather than selecting points directly. 
On the other hand, there is an increasing body of work designing neural network-based local feature aggregation operators for point clouds. Although some of them (\eg, PointCNN \cite{Li2018PointCNNCO}, PointASNL \cite{Yan2020PointASNLRP}, GSS \cite{Yang2019ModelingPC}) decrease the point number while learning latent features, they can hardly be considered as sampling methods in the true sense as no real spatial points exist during the processing. Moreover, none of the above methods consider shape outlines as special features.

In this paper, we propose \emph{a point cloud edge sampling method that combines neural network-based learning and mathematical statistics-based direct point selecting.} 
One key to the success of 2D image processing with neural networks is that they can detect primary edges and use them to form shape contours implicitly in the latent space \cite{Yosinski2015UnderstandingNN}. Inspired by that insight, we pursue the idea of focusing on salient outline points (edge points) for the sampling of point cloud subsets for downstream tasks.
Broadly speaking, edge detection may be considered a special sampling strategy. 
Hence, by revisiting the Canny edge detection algorithm \cite{Canny1986ACA} which is a widely-recognized classical edge detection method for images, we propose our attention-based point cloud edge sampling (APES) method for point clouds. It uses the attention mechanism \cite{vaswani2017attention} to compute correlation maps and sample edge points whose properties are reflected in these correlation maps.
We propose two kinds of APES with two different attention modes. Based on neighbor-to-point (N2P) attention which computes correlation maps between each point and its neighbors, local-based APES is proposed. Based on point-to-point (P2P) attention which computes a correlation map between all points, global-based APES is proposed. 
Our proposed method selects sampled points directly, and the intermediate result preserves the point index meaning so they can be visualized easily. Moreover, our method can downsample the input point cloud to any desired size.

We summarize our contributions as follows:
\begin{itemize}[itemsep=-3pt,topsep=-5pt,left=3pt]
\item A point cloud edge sampling method termed APES that combines neural network-based learning and mathematical statistics-based direct point selecting. 
\item Two variants of local-based APES and global-based APES, by using two different attention modes.
\item Good qualitative and quantitative results on common point cloud benchmarks, demonstrating the effectiveness of the proposed sampling method.
\end{itemize}

\section{Related Work}
\label{sec:relatedWork}
\subsection{Point Cloud Sampling}
In the past decades, non-learning-based sampling methods are mostly used for point cloud sampling.
FPS \cite{Eldar1994TheFP, Moenning2003FastMF} is the most widely used sampling method, which selects the farthest points iteratively. FPS is easy to implement and has been frequently used in neural networks that aggregate local features, \eg, PointNet++ \cite{Qi2017PointNetDH}, PointCNN \cite{Li2018PointCNNCO}, PointConv \cite{Wu2019PointConvDC}, and RS-CNN \cite{Liu2019RelationShapeCN}. Besides, RS has also been adopted to process large-scale point clouds with great computational efficiency in lots of works, including VoxelNet \cite{Zhou2018VoxelNetEL}, RandLA-Net \cite{Hu2020RandLANetES} and P2B \cite{Qi2020P2BPN}. A more recently proposed method of IDIS \cite{Groh2018FlexConvolutionM} defines the inverse density importance of a point by simply adding up all distances between the center point and its neighbors, and samples points whose sum values are smaller.

Recently, learning-based sampling methods show better performances on point cloud sampling when trained in a task-oriented manner. The pioneering work of S-Net \cite{Dovrat2019LearningTS} generates new point coordinates directly from the global representation. Its subsequent work of SampleNet \cite{Lang2020SampleNetDP} further introduces a soft projection operation for better point approximation in the post-processing step. Alternatively, DA-Net \cite{Lin2021DANetDD} extends S-Net with a density-adaptive sampling strategy, which decreases the influence of noisy points.
By learning a sampling transformation matrix, MOPS-Net \cite{Qian2020MOPSNetAM} multiplies it with the original point cloud to generate a new one as the sampled point cloud. 
CPL \cite{Nezhadarya2020AdaptiveHD} samples points by investigating the output in the max-pooling layer.
Replacing the MLP layers in S-Net with several self-attention layers, PST-NET \cite{Wang2021PSTNETPC} reports better performances on trained tasks. Its subsequent work of LighTN \cite{Wang2022LighTNLT} proposes a lightweight Transformer framework for resource-limited cases.

\subsection{Deep Learning on Point Clouds}
Prior to the emergence of PointNet \cite{Qi2017PointNetDL}, deep learning-based methods for point cloud analysis are usually multi-view-based \cite{lawin2017deep, boulch2017unstructured, audebert2016semantic, tatarchenko2018tangent} or volumetric-based \cite{maturana2015voxnet, jiang2018pointsift, le2018pointgrid}. 
PointNet \cite{Qi2017PointNetDL} is the first DL-based method that learns directly on points and it uses point-wise MLP to extract global features. Its subsequent work of PointNet++ \cite{Qi2017PointNetDH} further considers local information. Convolution-based methods \cite{Zhao2019PointWeb, Wu2019PointConvDC, Thomas2019KPConvFA, Li2018PointCNNCO, Lin2020FPConv, Xu2021PAConvPA, Wiersma2021DeltaConvAP} bring the convolution operation into point cloud feature learning. For example, PointConv\cite{Wu2019PointConvDC} and KPConv \cite{Thomas2019KPConvFA} propose point-wise convolution operators with which points are convoluted with neighbor points. 
Graph-based methods \cite{Wang2019DynamicGC, Liu2019RelationShapeCN, Zhang2021LinkedDG, Xu2020GridGCN, Chen2021GAPNetGA, Liang20203DIE, Lin2020Convolution} analyze point clouds by using graph structure. For example, Simonovsky \etal \cite{simonovsky2017dynamic} treat each point as a graph vertex and apply graph convolution. In DGCNN \cite{Wang2019DynamicGC}, EdgeConv blocks update the neighbor information dynamically based on dynamic graphs. 
More recently, Attention-based methods \cite{Guo2021PCTPC, Engel2021PointT, Hu2020RandLANetES, Pan20213DOD, Bhattacharyya2021SADet3DSB, Cheng2021PatchFormerAV, Lu20223DCTN3C, Lu20223DPCT3P} are starting to trend. PCT \cite{Guo2021PCTPC} pioneers this direction by replacing the encoder layers in the PointNet framework with self-attention layers, while PT \cite{Zhao2021PointT} is based on U-Net \cite{Ronneberger2015UNetCN}. 3DCTN \cite{Lu20223DCTN3C} uses offset attention blocks, while a deformable self-attention module is proposed in SA-Det3D \cite{Bhattacharyya2021SADet3DSB}, and a dual self-attention module is proposed in 3DPCT \cite{Lu20223DPCT3P}. 
Stratified Transformer \cite{Lai2022StratifiedTF} additionally samples distant points as the key input to capture long-range contexts.

\begin{figure}[t]
    \centering
    \includegraphics[width=1\linewidth,trim=0 0 0 0,clip]{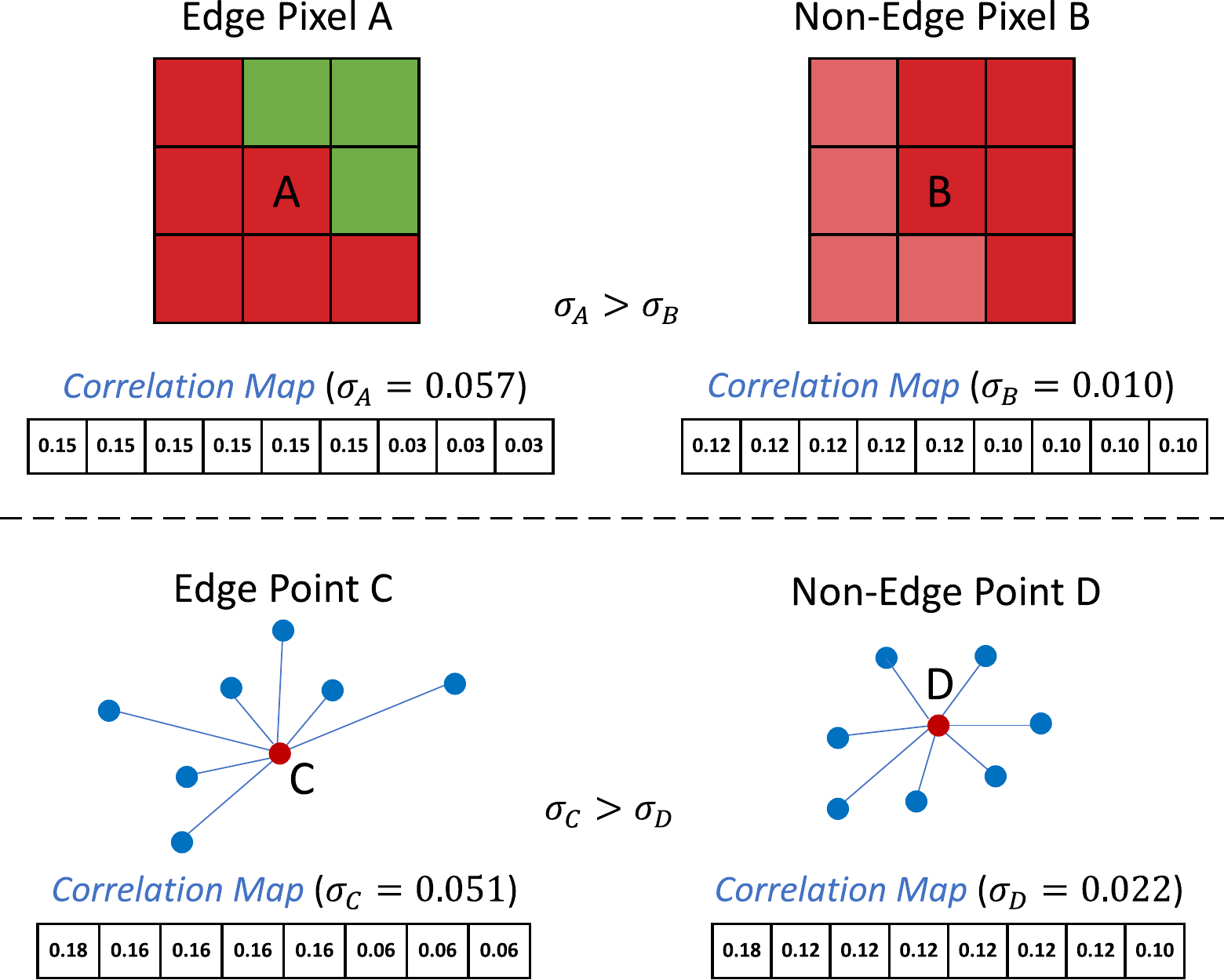}
    \caption{Illustration of using standard deviation to select edge pixels/points. A normalized correlation map is computed between the center pixel/point and its neighbors. The center pixel/point is self-contained as a neighbor. A larger standard deviation in the normalized correlation map means a higher possibility that it is an edge pixel/point. \vspace{-0.2cm}}
    \label{fig:ds_sigma}
\end{figure}

\section{Methodology}
\label{sec:method}
\subsection{Revisiting Canny Edge Detection on Images}
The Canny edge detector uses a multi-stage algorithm to detect edges in images. It consists of five steps: (i) Apply Gaussian filter to smooth the image; (ii) Find the intensity gradients of the image; (iii) Apply gradient magnitude thresholding or lower bound cut-off suppression; (iv) Apply double threshold to determine potential edges; (v) Finalize the detection of edges by suppressing all the other edges that are weak and not connected to strong edges. 

The key to the effectiveness of the Canny edge detector is how edge pixels are defined. 
The intensity gradient of each pixel $i$ is computed in comparison to its neighbors in a patch set $\cS_i$, which is typically a $3 \times 3$ or $5 \times 5$ patch. Pixels with larger intensity gradients are defined as edge pixels. We make the following observation: 
\emph{If there are large differences between the pixels from a patch set $\cS_i$, then the standard deviation $\sigma_i$ of the intensities in the patch is also high.}
Hence, an alternative method for edge detection is to select pixels whose patch sets have larger $\sigma_i$.

We further generalize beyond pixel intensities to any (latent) per-pixel features $\bp_i$ with a \enquote{measure of feature correlation} $h(\bp_i, \bp_{ij})$ defined between the center pixel $i$ and its neighbor pixel $j$. 
In each patch $\cS_i$, we call the vector $\bm_i = \mathrm{softmax}\big(h(\bp_i, \bp_{ij})_{j\in \cS_i}\big)$ the normalized correlation map between the center pixel and its neighbors.
Then the standard deviation $\sigma_i$ is computed over the elements of $\bm_i$, and \emph{pixels with larger $\sigma_i$ are selected as edge pixels.}
An illustration is given in the top row of Figure \ref{fig:ds_sigma}. When the neighbor number $k$ is fixed (\eg, $k=9$ for the top row, the center pixel is self-contained as a neighbor), for each patch, the mean value of its normalized correlation map is always $1/k$. However, for edge pixels, the standard deviations of their normalized correlation maps are larger.

For images, the proposed alternative edge detection algorithm, and in particular using the standard deviation for the normalized correlation map, is computationally much more expensive compared to the Canny edge detector. However, it provides the starting point to transfer the idea to point cloud edge sampling.
Unlike images where pixels are well-aligned and patch operators can be easily defined and applied, point clouds are usually irregular, unordered, and potentially sparse. Voxel-based 3D convolution kernels are not applicable. Moreover, image pixels come with a color value (\eg RGB or grayscale). For many point clouds, however, the point coordinates are the only available feature.

\begin{figure}[t]
    \centering
    \includegraphics[width=0.9\linewidth,trim=0 0 0 0,clip]{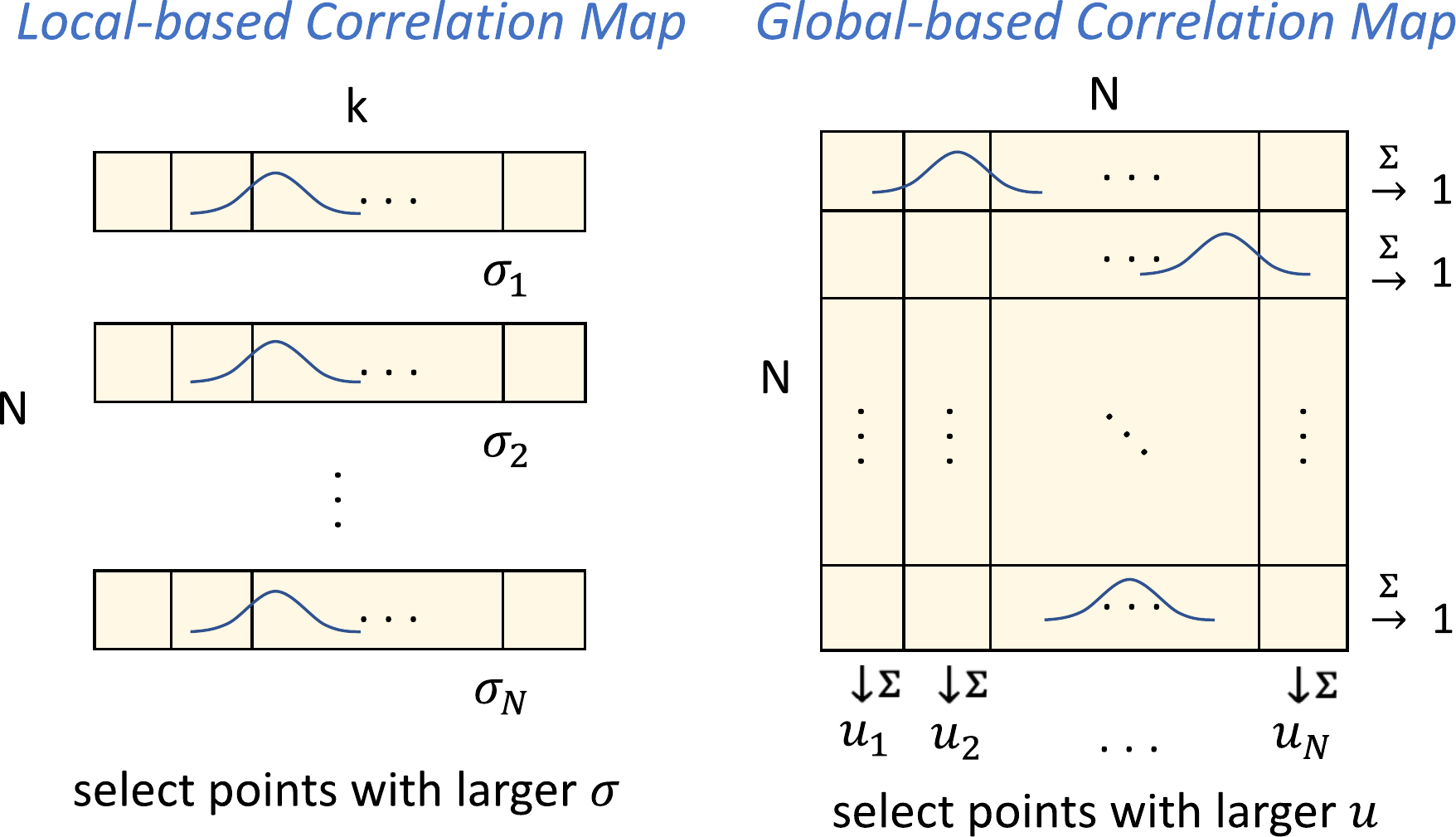}
    \caption{The key idea of proposed methods. $N$ denotes the total number of points, while $k$ denotes the number of neighbors used for local-based sampling method. \vspace{-0.2cm}}
    \label{fig:methods}
\end{figure}

\begin{figure*}[t]
    \centering
    \includegraphics[width=1\linewidth,trim=2 2 2 2,clip]{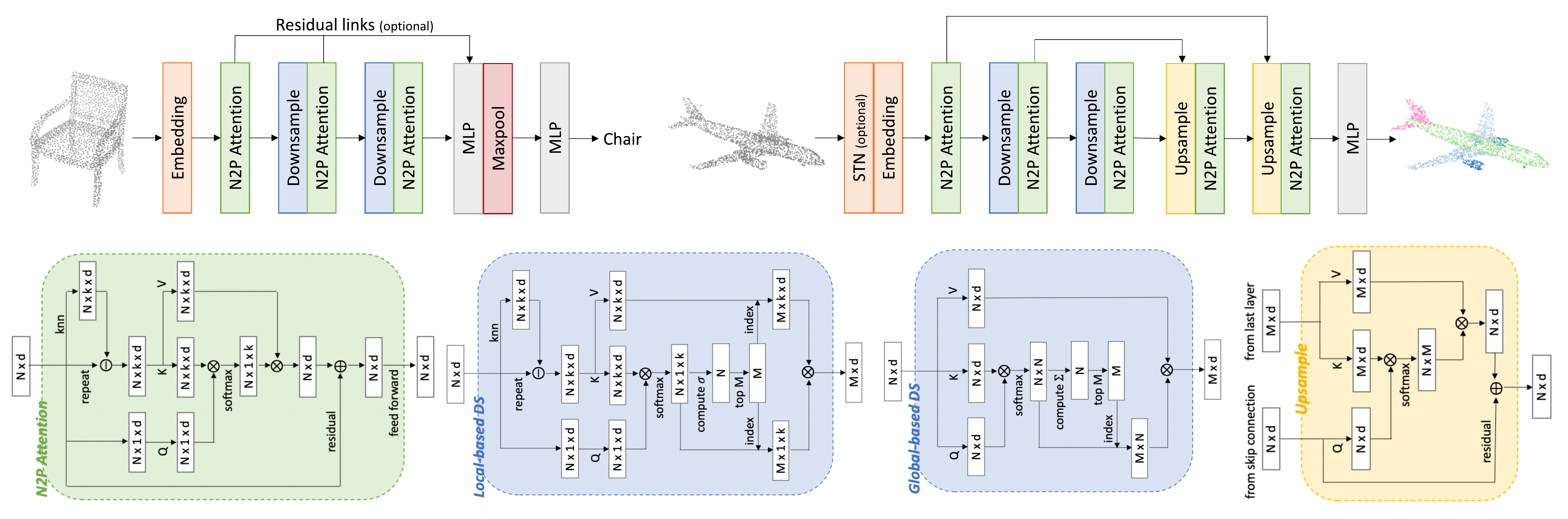}
    \caption{Network architectures for classification (top left) and segmentation (top right). The structures of N2P attention feature learning layer (bottom left), two alternative downsample layers (bottom middle), and upsample layer (bottom right) are also given. Both kinds of downsample layers downsample a point cloud from $N$ points to $M$ points, while upsample layer upsamples it from $M$ points to $N$ points.}
    \label{fig:net}
\end{figure*}

\subsection{Local-based Point Cloud Edge Sampling}
To adopt the previously introduced alternative edge detection algorithm to a point cloud set with $|\cS|=N$ points, we use $k$-nearest neighbor to define a local patch $\cS_i\subseteq \cS$ for each point $i$ to compute normalized correlation maps. 
As illustrated in the bottom row of Figure \ref{fig:ds_sigma}, when the neighbor number $k$ is fixed (\eg, $k=8$ for the bottom row, the center point is self-contained as a neighbor), for each patch, the mean value of its normalized correlation map is again always $1/k$. However, for edge points, the standard deviations of their normalized correlation maps are larger.

On the other hand, the attention mechanism is an ideal option to serve as the \enquote{measure of correlation} between point features within each patch, \ie, \emph{the attention map serves as the normalized correlation map} directly. The local-based correlation measure $h^l(\cdot)$ is defined as
\begin{equation}
h^l(\bp_i, \bp_{ij}) = Q(\bp_i)^{\top} K(\bp_{ij} - \bp_i)
\end{equation}
where $Q$ and $K$ stand for the linear layers applied on the query input and the key input, respectively. Here we use the (latent) features of the center point $\bp_i$ as the query input, and the feature difference between the neighbor point and the center point $\bp_{ij} - \bp_i$ as the key input. 
As in the original Transformer model \cite{vaswani2017attention}, the square root of the feature dimension count  $\sqrt{d}$ serves as a scaling factor.
The final normalized correlation map $\bm^l_i$ is given as
\begin{equation}
\bm^l_i = \mathrm{softmax}\left(h^l(\bp_i, \bp_{ij})_{j\in \cS_i}/\sqrt{d}\right)\,.
\end{equation}
Again, a standard deviation $\sigma_i$ is computed for each normalized correlation map. \emph{The edge points are sampled by selecting the points with higher $\sigma_i$.}

\subsection{Global-based Point Cloud Edge Sampling}
We term the above-applied attention as neighbor-to-point (N2P) attention, which is specifically designed to capture local information using patches. For sampling problems, global information is also crucial. 
Considering the special case where all points are included in the local patch (\ie, $k=N$), a new global correlation map $M^g$ of size $N \times N$ is obtained with the linear layers $Q$ and $K$ being shared for all points. Now the N2P attention simplifies into the common self-attention. We term it point-to-point (P2P) attention in this paper.
In this case, the correlation measure $h^g(\cdot)$ and the normalized correlation map are defined as:
\begin{gather}
h^g(\bp_i, \bp_j) = Q(\bp_i)^{\top} K(\bp_j)\\
\bm^g_i = \mathrm{softmax}\left(h^g(\bp_i, \bp_{j})_{j\in \cS}/\sqrt{d}\right)
\end{gather}
Note that all $\bm^g_i$ now have the same point order, but the attention result for each point pair is not affected by the order.

The global correlation map $M^g$ regroups the point-wise normalized correlation maps into a $N\times N$ matrix:
\begin{equation}
M^g = 
\begin{pmatrix}
\rule[.5ex]{1.5em}{0.4pt} \hspace{-0.6em} & {\bm^g_{1}}^{\top} & \hspace{-0.6em} \rule[.5ex]{1.5em}{0.4pt} \\
\rule[.5ex]{1.5em}{0.4pt} \hspace{-0.6em} & {\bm^g_{2}}^{\top} & \hspace{-0.6em} \rule[.5ex]{1.5em}{0.4pt} \\
  & \vdots  &   \\
\rule[.5ex]{1.5em}{0.4pt} \hspace{-0.6em} & {\bm^g_{N}}^{\!\!\top} & \hspace{-0.6em} \rule[.5ex]{1.5em}{0.4pt}
\end{pmatrix}
\end{equation}
In the context of the global correlation map $M^g$, instead of computing row-wise standard deviations for selecting points, we propose an alternative approach. 
Denote $m_{ij}$ as the value of $i$th row and $j$th column in $M^g$. For point $i$, if it is an edge point, $\bm^g_i$ has a larger standard deviation. In this case, considering its neighboring area, if a point $j$ is close (based on 3d spatial space or latent space) to point $i$, $m_{ij}$ is larger and point $j$ is also likely to be an edge point. Given this property, now consider $M^g$ column-wise. For a point $j$, in order to qualify it as an edge point, it needs to rank a higher value of $m_{ij}$ in $M^g$ more often compared to other points.
Hence \emph{instead of computing row-wise standard deviations, we compute column-wise sums}. 
Denote $u_j = \sum_i m_{ij}$, we then \emph{sample the points with higher $u_j$.}
Overall, we can consider it as follows: if a point contributes more in the self-attention correlation map, it is more likely to be an \enquote{important} point.
An illustrative figure of the two proposed methods is given as Figure \ref{fig:methods}. 

\begin{figure*}[t]
    \centering
    \includegraphics[width=0.95\linewidth,trim=0 2 0 0,clip]{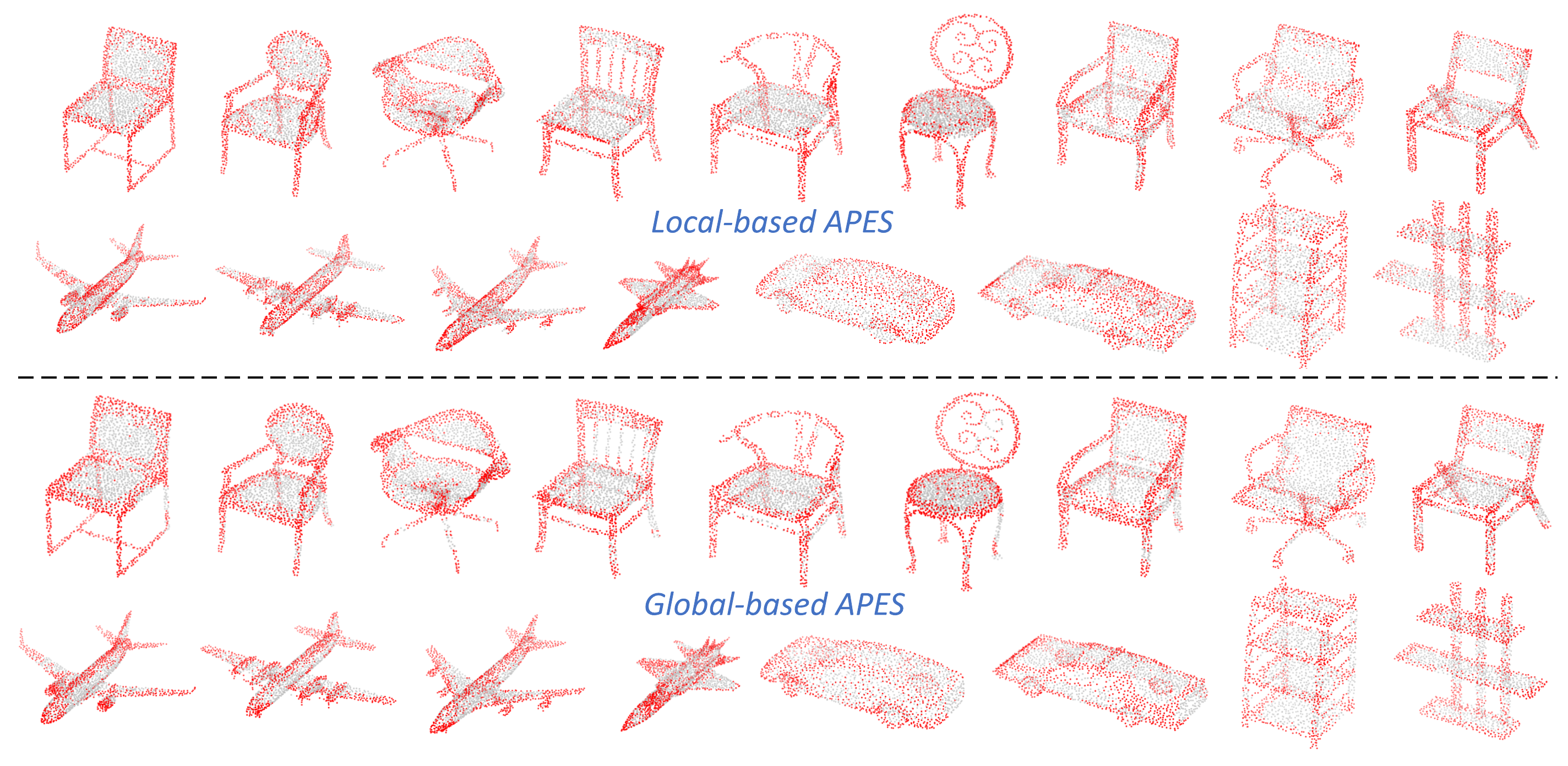}
    \caption{Visualized sampling results of local-based APES and global-based APES on different shapes. All shapes are from the test set.}
    \label{fig:vis_cls}
\end{figure*}

\section{Experimental Results}
\label{sec:experiments}

\subsection{Classification}
\label{sec:cls}
\textbf{Dataset.}
ModelNet40 \cite{Wu20153DSA} contains 12311 manufactured 3D CAD models in 40 common object categories. For a fair comparison, we use the official train-test split, in which 9843 models are used for training and 2468 models for testing. From each model mesh surface, points are uniformly sampled and normalized to the unit sphere. Only 3D coordinates are used as point cloud input. For data augmentation, we randomly scale, rotate, and shift each object point cloud in the 3D space.
During the test, no data augmentation or voting methods were used.

\textbf{Network Architecture Design.}
The classification network architecture is given in Figure \ref{fig:net}.
The embedding layer converts the input 3D coordinates into a higher-dimensional feature with multiple EdgeConv blocks.
For feature learning layers, it is possible to use the layers designed for a similar purpose in other papers.
Alternatively, the aforementioned N2P attention or P2P attention can also be used as feature learning layers. 
We use $k=32$ neighbor points as default in local-based APES downsample layers. For an input point cloud of $N$ points from the previous layer, each downsample layer samples it to $N/2$ points. Note that our method can actually sample the point cloud to any desired number of points. The optional residual links are used for better feature transmission.

\textbf{Training Details.}
To train the model, we use AdamW optimizer with an initial
learning rate $1\times 10^{-4}$ and decay it to $1\times 10^{-8}$ with a cosine annealing schedule. 
The weight decay hyperparameter for network weights is set as $1$. Dropout with a probability of $0.5$ is used in the last two fully connected layers. 
We train the network with a batch size of 8 for 200 epochs.

\textbf{Quantitative and Qualitative Results.}
The quantitative comparison with the SOTA methods is summarized in Table \ref{table:cls}, where our proposed APES is among the best-performing methods. Qualitative results are presented in Figure \ref{fig:vis_cls}. From it, we can observe that both local-based APES and global-based APES achieve good edge sampling results. On the other hand, local-based APES focuses more strictly on edge points, while global-based APES ignores some edge points and leverages a bit more on other non-edge points that are close to the edges. 
For example, in chair shapes, global-based APES discards some chair leg points and selects more points for chair seat edges to make the edges “thicker”. We contribute its slightly better quantitative results to this.
Overall, sampling more edge points improves the performance of downstream tasks. However, this can be overdone, and selecting only edge points can be detrimental. APES uses end-to-end training that includes the downstream task to make a good trade-off in the sample selection.
Local-based APES imposes stronger mathematical statistics constraints during the task loss minimization, while global-based APES pursues better performance by allowing sampling the points that are less belong to the edge yet more important globally.

\begin{table}[t]
\centering
\scalebox{0.82}{
\begin{tabular}{lc}
\toprule
Method & Overall Accuracy \\ \midrule
PointNet \cite{Qi2017PointNetDL} & 89.2\% \\
PointNet++ \cite{Qi2017PointNetDH} & 91.9\% \\
SpiderCNN \cite{Xu2018SpiderCNNDL} & 92.4\% \\
DGCNN \cite{Wang2019DynamicGC} & 92.9\% \\
PointCNN \cite{Li2018PointCNNCO} & 92.2\% \\
PointConv \cite{Wu2019PointConvDC} & 92.5\% \\
PVCNN \cite{Liu2019PointVoxelCF} & 92.4\% \\
KPConv \cite{Thomas2019KPConvFA} & 92.9\% \\
PointASNL \cite{Yan2020PointASNLRP} & 93.2\% \\
PT$^1$ \cite{Engel2021PointT} & 92.8\% \\
PT$^2$ \cite{Zhao2021PointT} & 93.7\% \\
PCT \cite{Guo2021PCTPC} & 93.2\% \\
PRA-Net \cite{Cheng2021PRANetPR} & 93.7\% \\
PAConv \cite{Xu2021PAConvPA}  & 93.6\% \\
CurveNet \cite{Muzahid2021CurveNetCM} & \textbf{93.8\%} \\
DeltaConv \cite{Wiersma2021DeltaConvAP} & \textbf{93.8\%} \\ \midrule
APES (local-based) & 93.5\% \\
APES (global-based) & \textbf{93.8\%} \\ \bottomrule
\end{tabular}}
\caption{Classification results on ModelNet40. In comparison with other SOTA methods that also only use raw point clouds as input. Note that our reported results did not consider the voting strategy.}
\label{table:cls}
\end{table}

\begin{figure*}[t]
    \centering
    \includegraphics[width=0.95\linewidth,trim=0 0 0 0,clip]{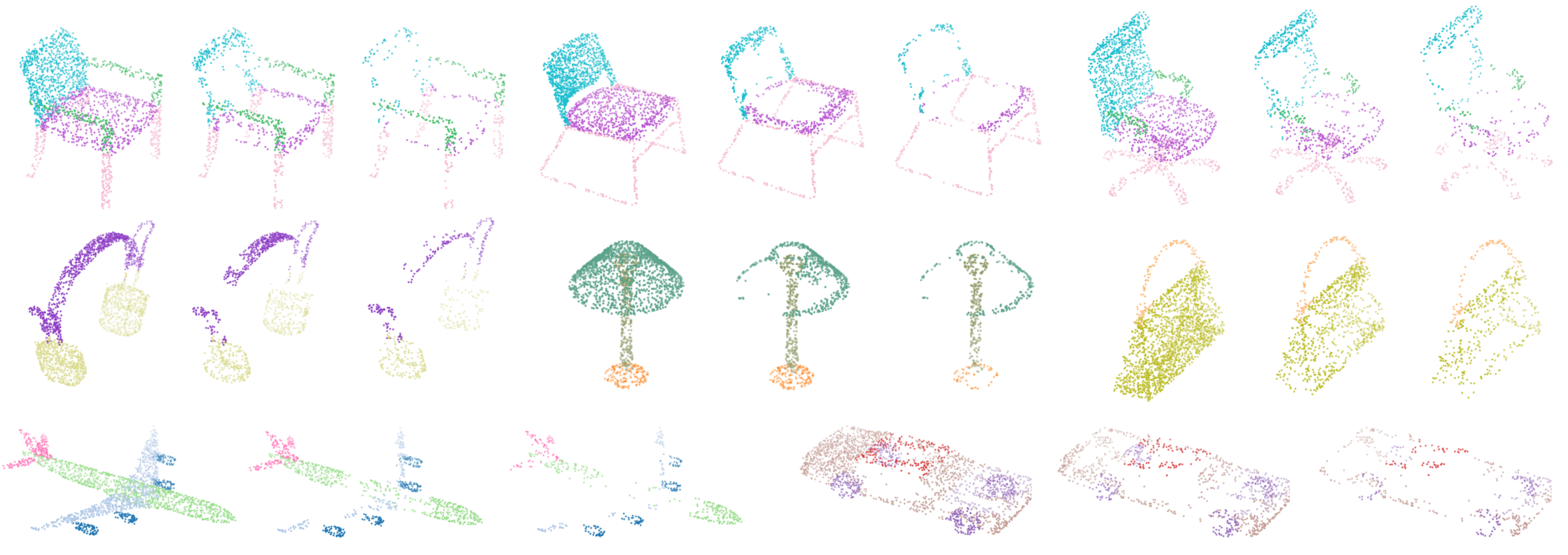}
    \caption{Visualized segmentation results as shape point clouds are downsampled. All shapes are from the test set.}
    \label{fig:vis_seg}
\end{figure*}

\subsection{Part Segmentation}
\label{sec:seg}
\textbf{Dataset.}
The ShapeNetPart dataset \cite{Yi2016ASA} is annotated for 3D object part segmentation. It consists of 16,880 models from 16 shape categories, with 14,006 3D models for training and 2,874 for testing. The number of parts for each category is between 2 and 6, with 50 different parts in total. We use the sampled point sets produced in \cite{Qi2017PointNetDH} for a fair comparison with prior work. For evaluation metrics, we report category mIoU and instance mIoU.

\textbf{Network Architecture Design.}
The segmentation network architecture is given in Figure \ref{fig:net}.
Most network layers are identical to the layers in the classification model, except for the spatial transform network (STN) and the upsample layer. 
The optional STN layer learns a spatial transformation matrix to transform the input cloud for better initial alignment \cite{Qi2017PointNetDL, Wang2019DynamicGC}. 
The upsample layer is a cross-attention-based layer to map the input point cloud to an upsampled size. Its key and value input is the feature from the last layer, while the query input is the corresponding residual feature within the downsample process.

\textbf{Training Details.}
To train the model, we use AdamW optimizer with an initial
learning rate $1\times 10^{-4}$ and decay it to $1\times 10^{-8}$ with a cosine annealing schedule. The weight decay hyperparameter for network weights is set as $1\times 10^{-4}$. The dropout with a probability of $0.5$ is used in the last two fully connected layers. 
We train the network with a batch size of 16 for 200 epochs.

\begin{table}[t]
\centering
\scalebox{0.82}{
\begin{tabular}{lcc}
\toprule
Method & Cat. mIoU & Ins. mIoU \\ \midrule
PointNet \cite{Qi2017PointNetDL} & 80.4\% & 83.7\% \\
PointNet++ \cite{Qi2017PointNetDH} & 81.9\% & 85.1\% \\
SpiderCNN \cite{Xu2018SpiderCNNDL} & 82.4\% & 85.3\% \\
DGCNN \cite{Wang2019DynamicGC} & 82.3\% & 85.2\% \\
SPLATNet \cite{Su2018SPLATNetSL} & 83.7\% & 85.4\% \\
PointCNN \cite{Li2018PointCNNCO} & 84.6\% & 86.1\% \\
PointConv \cite{Wu2019PointConvDC} & 82.8\% & 85.7\% \\
KPConv \cite{Thomas2019KPConvFA} & 85.0\% & 86.2\% \\
PT$^1$ \cite{Engel2021PointT} & - & 85.9\% \\
PT$^2$ \cite{Zhao2021PointT} & 83.7\% & \textbf{86.6\%} \\
PCT \cite{Guo2021PCTPC} & - & 86.4\% \\
PRA-Net \cite{Cheng2021PRANetPR} & 83.7\% & 86.3\% \\
PAConv \cite{Xu2021PAConvPA} & 84.6\% & 86.1\% \\
CurveNet \cite{Muzahid2021CurveNetCM} & - & \textbf{86.6\%} \\
StratifiedTransformer \cite{Lai2022StratifiedTF} & \textbf{85.1\%} & \textbf{86.6\%} \\
\midrule
APES (local-based) & 83.1\% & 85.6\% \\
APES (global-based) & 83.7\% & 85.8\% \\ \bottomrule
\end{tabular}}
\caption{Segmentation results on ShapeNet Part.}
\label{table:seg}
\end{table}

\begin{table}[ht]
\centering
\resizebox{1\linewidth}{!}{
\begin{tabular}{lccccccc}
\toprule
\multirow{2}{*}{\diagbox{Method}{Points}} & \multicolumn{3}{c}{Cat. mIoU (\%)} & \multirow{2}{*}{} & \multicolumn{3}{c}{Ins. mIoU (\%)} \\ \cmidrule{2-4} \cmidrule{6-8} 
 & 2048 & 1024 & 512 &  & 2048 & 1024 & 512 \\ \midrule
APES (local) & 83.11 & 85.56 & \textbf{86.17} &  & 85.58 & 87.27 & 87.41  \\
APES (global) & 83.67 & 84.86 & 85.44 &  & 85.81 & 87.78 & \textbf{88.06} \\ \bottomrule
\end{tabular}}
\caption{Segmentation results of the full point clouds and intermediate downsampled point clouds of different sizes.}
\label{table:seg_ds}
\end{table}

\textbf{Quantitative and Qualitative Results.}
The segmentation quantitative results are given in Table \ref{table:seg}. Our method achieves decent performance but is not on par with the best ones. However, as we compute the same metrics on the intermediate downsampled point clouds in Table \ref{table:seg_ds}, we surprisingly find that their performances are extremely good, even far better than the SOTA methods. This indicates the downsampled edge points contribute more to the performance, while the features of the discarded non-edge points are not well reconstructed by the upsample layer.
Most other neural network papers use FPS for downsampling and FPS preserves a similar data distribution compared to the original point cloud. When upsampling, simple interpolation operations \cite{Qi2017PointNetDH, Zhao2021PointT, Lai2022StratifiedTF} are used to create new points.
However, our method focuses on edge points and the sampled result has a quite different data distribution than the original point cloud. 
For non-edge points, especially those far from edges, neighbor-based interpolation methods are no longer applicable.
We have designed a cross attention-based layer for upsampling, but it is still hard to get the features of the former discarded points back, even with residual links. 
Note that in this case, the upsampling problem actually turns into a point cloud completion or reconstruction task, which is another advanced topic for point cloud analysis. We would like to leave this for future work. The qualitative segmentation results are given in Figure \ref{fig:vis_seg}, intermediate visualization 
results are also provided.

\begin{table}[t]
\centering
\resizebox{0.85\linewidth}{!}{
\begin{tabular}{ccc}
\toprule
Method & Feature Learning Layer & OA (\%) \\ \midrule
DGCNN & EdgeConv & 92.90 \\ \midrule
\multirow{3}{*}{APES (local-based)} & EdgeConv & 93.02 \\
 & P2P Attention & 93.30 \\
 & N2P Attention & \textbf{93.47} \\ \midrule
\multirow{3}{*}{APES (global-based)} & EdgeConv & 93.18 \\
 & P2P Attention & 93.46 \\
 & N2P Attention & \textbf{93.81} \\ \bottomrule
\end{tabular}}
\caption{Ablation study of using different feature learning layers in the classification network.}
\label{table:FL_layer}
\end{table}

\subsection{Ablation study}
\label{sec:ablation}
In this subsection, multiple ablation studies are conducted regarding the design choices of neural network architectures. 
All following experiments are performed on the classification benchmark of ModelNet40.

\textbf{Feature Learning Layer.}
The feature learning layer we used in the above experiments is the N2P attention layer. However, as discussed in Section \ref{sec:cls}, it is possible to replace it with other feature layers. We additionally report the results of using EdgeConv or P2P attention as the feature learning layer in Table \ref{table:FL_layer}. From it, we can observe that N2P attention achieves the best performance. Meanwhile, the results of using EdgeConv are improved when using our proposed sampling methods.

\begin{table}[t]
\centering
\resizebox{0.85\linewidth}{!}{
\begin{tabular}{ccc}
\toprule
Method & Embedding Dimension & OA (\%) \\ \midrule
\multirow{3}{*}{APES (local-based)} & 64 & 93.10 \\
 & 128 & 93.47 \\
 & 192 & \textbf{93.54} \\ \midrule
\multirow{3}{*}{APES (global-based)} & 64 & 93.34 \\
 & 128 & 93.81 \\
 & 192 & \textbf{93.83} \\ \bottomrule
\end{tabular}}
\caption{Ablation study of using a different number of embedding dimensions for the classification task. \vspace{-0.2cm}}
\label{table:embedding}
\end{table}

\textbf{Embedding Dimension.}
In most network-based methods, it is often reported that better performances are achieved when a larger embedding dimension is used. In our experiments, we use an embedding dimension of 128 as the default. We additionally report the results of using embedding dimensions of 64 and 192 in Table \ref{table:embedding}.

\textbf{Choice of $k$ in local-based APES.}
When local-based APES is used, the parameter of neighbor number $k$ is a very important parameter since it decides the perception area size of local patches. We additionally report the results of using different $k$ in Table \ref{table:local_k}.

\begin{table}[t]
\centering
\resizebox{1\linewidth}{!}{
\begin{tabular}{cccccccc}
\toprule
$k$ & 8 & 16 & 32 & 64 & 128 & 256 & 512 \\ \midrule
OA (\%) & 93.14 & 93.26 & 93.47 & 93.52 & 93.54 & 93.59 & \textbf{93.63} \\ \bottomrule
\end{tabular}}
\caption{Ablation study of using a different number of neighbors for local-based edge point sampling.}
\label{table:local_k}
\end{table}

\begin{table}[t]
\centering
\resizebox{1.0\linewidth}{!}{
\begin{tabular}{cccc}
\toprule
Edge Supervision & None & Pre-trained and Fixed & Joint Training \\ \midrule
APES (local-based) & 93.47\% & 93.45\% & 93.46\% \\
APES (global-based) & \textbf{93.81\%} & 93.47\% & 93.51\% \\ \bottomrule
\end{tabular}}
\caption{Ablation study of considering the edge supervision. Results of using it for pre-training or joint training are both presented.}
\label{table:edgeSupervision}
\end{table}

\begin{figure}[t]
    \centering
    \includegraphics[width=1\linewidth,trim=0 0 0 0,clip]{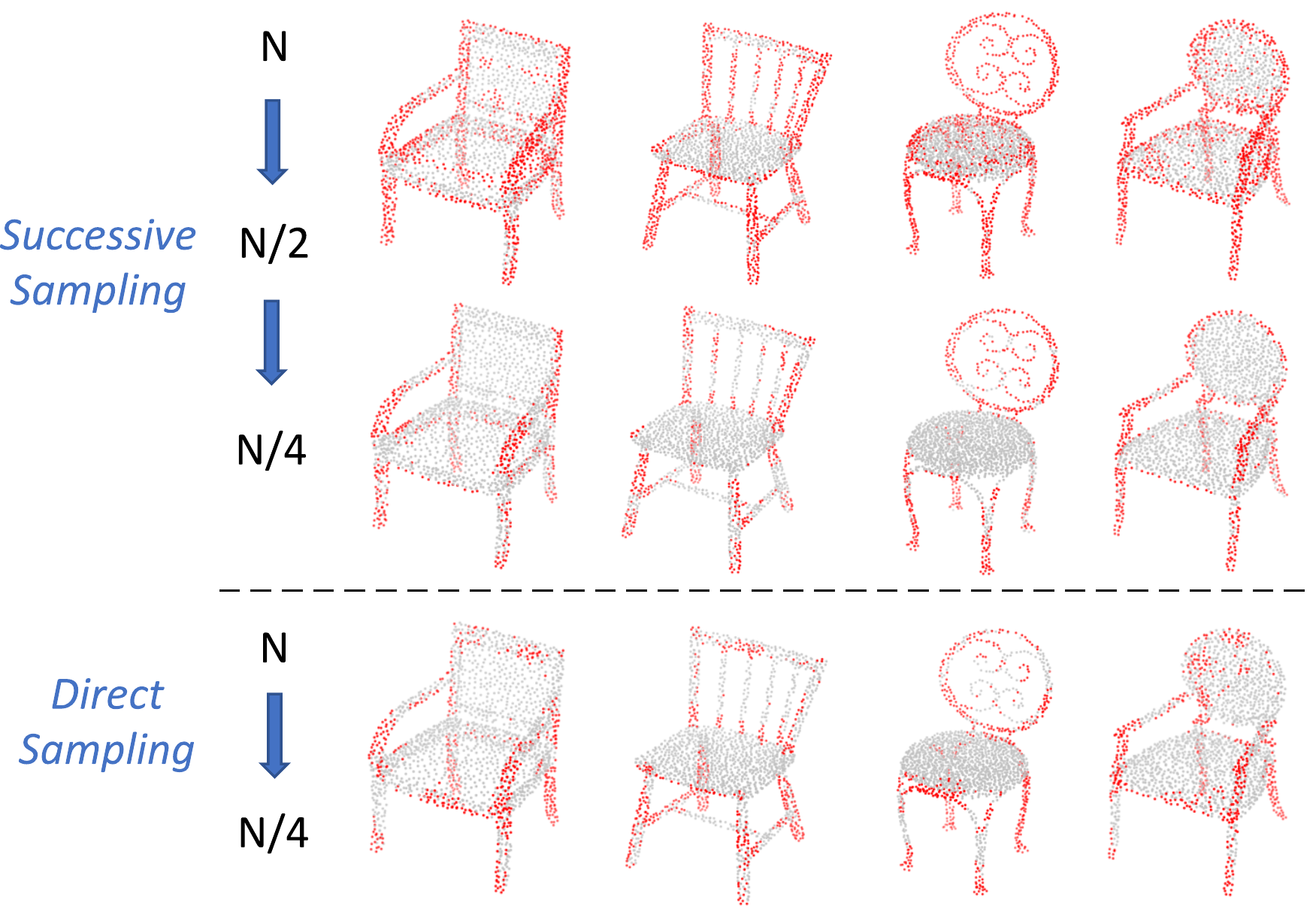} 
    \caption{Sampling results of successively sampling to a fourth of the original size and directly sampling by a factor of four. 
    }
    \label{fig:successive}
\end{figure}

\textbf{Successive sampling vs. Direct sampling.}
An advantage of our proposed method is that we can sample any desired number of points with it.
We further provide qualitative comparison results of successively sampling the raw point cloud to a quarter ($N\rightarrow N/2 \rightarrow N/4$) and directly sampling it to a quarter ($N\rightarrow N/4$) in Figure \ref{fig:successive}. We observe that the sampled results are mostly similar. With the exception of a few extreme edge points which are better captured by successive sampling.

\textbf{Additional edge point supervision.} Since it is possible to compute "ground-truth" edge points from the shapes using local curvatures, we further study the cases where an edge supervision loss term is introduced. Experiments of not using the edge supervision, using it for pre-training (and fixing it during the downstream task training), and using it for joint training are conducted.
Numerical results are given in Table \ref{table:edgeSupervision}. The results are consistent with our conclusion in subsection \ref{sec:cls}. For local-based APES which already focuses on edge point sampling, edge supervision has no significant impact. However, for global-based APES, edge supervision decreases performance slightly.

\begin{table*}[t]
\centering
\resizebox{1\linewidth}{!}{
\begin{tabular}{c|ccccccccccc}
\toprule
$M$ & Voxel & RS & FPS \cite{Eldar1994TheFP} & S-NET \cite{Dovrat2019LearningTS} & PST-NET \cite{Wang2021PSTNETPC} & SampleNet \cite{Lang2020SampleNetDP} & MOPS-Net \cite{Qian2020MOPSNetAM} & DA-Net \cite{Lin2021DANetDD} & LighTN \cite{Wang2022LighTNLT} & APES (local) & APES (global) \\ \midrule
512 & 73.82 & 87.52 & 88.34 & 87.80 & 87.94 & 88.16 & 86.67 & 89.01 & 89.91 & 90.79 & \textbf{90.81}\\
256 & 73.50 & 77.09 & 83.64 & 82.38 & 83.15 & 84.27 & 86.63 & 86.24 & 88.21 & 90.38 & \textbf{90.40} \\
128 & 68.15 & 56.44 & 70.34 & 77.53 & 80.11 & 80.75 & 86.06 & 85.67 & 86.26 & 89.73 & \textbf{89.77} \\
64 & 58.31 & 31.69 & 46.42 & 70.45 & 76.06 & 79.86 & 85.25 & 85.55 & 86.51 & 88.68 & \textbf{89.57} \\
32 & 20.02 & 16.35 & 26.58 & 60.70 & 63.92 & 77.31 & 84.28 & 85.11 & 86.18 & 86.49 & \textbf{88.56} \\ 
\bottomrule
\end{tabular}}
\caption{Comparison with other sampling methods. Evaluated on the ModelNet40 classification benchmark with multiple sampling sizes. 
\vspace{-0.1cm}
}
\label{table:compare}
\end{table*}

\begin{table}[t]
\centering
\resizebox{1\linewidth}{!}{
\begin{tabular}{c|ccccccccccc}
\toprule
Method & S-NET & PST-NET & SampleNet & MOPS-Net & LighTN & APES (local) & APES (global) \\ \midrule
Params & 0.33M & 0.42M & 0.46M & 0.44M & 0.37M & 0.35M & 0.35M \\
FLOPs & 152M & 122M & 167M & 149M & 115M & 142M & 114M \\
\bottomrule
\end{tabular}}
\caption{Computation complexity of different sampling methods. Here \enquote{M} stands for million. \vspace{-0.1cm}} 
\label{table:time}
\end{table}

\begin{figure}[t]
    \centering
    \setlength{\abovecaptionskip}{2pt}
    \includegraphics[width=1\linewidth,trim=0 0 0 0,clip]{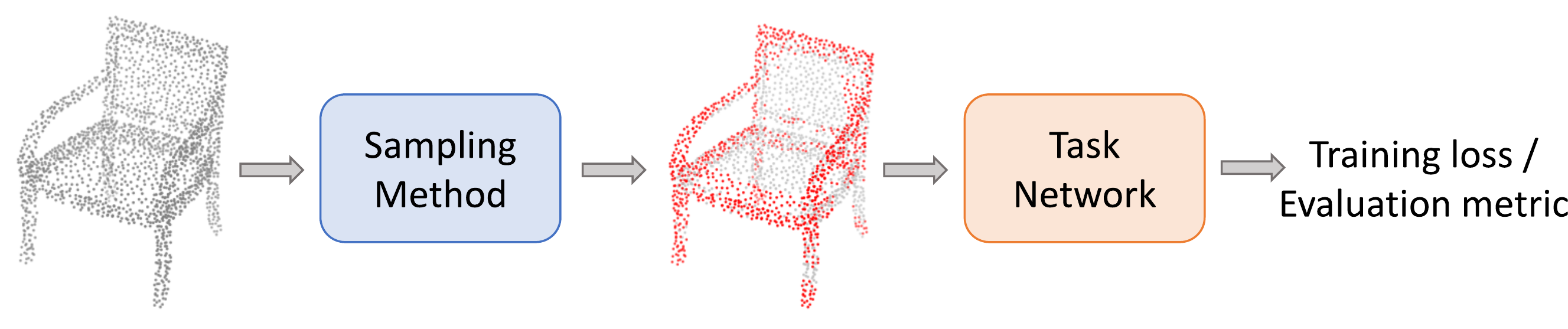} 
    \caption{Framework for sampling methods evaluation.
    \vspace{-0.2cm}
    }
    \label{fig:compare}
\end{figure}

\section{Sampling Methods Comparison}
\label{sec:compare}
\subsection{Experiment Setting}
We additionally compare our sampling method to previous work including RS, FPS, and the more recent learning-based S-Net, SampleNet, LighTN, etc. The same evaluation framework from \cite{Dovrat2019LearningTS, Lang2020SampleNetDP, Wang2022LighTNLT} is used, as illustrated in Figure \ref{fig:compare}. The task here is the ModelNet40 Classification, and the task network is PointNet. 
Sampling methods are evaluated with multiple sampling sizes. 

As discussed in the results part of Section \ref{sec:seg}, edge point sampling changes the data distribution compared to the original point cloud, especially when a large downsampling ratio (defined as $N/M$) is used. 
Hence for a fair comparison, in order to achieve a downsampled point cloud size of $M$, we first sample the input point cloud to a size of $2M$ with FPS, then sample it to the desired size $M$ with our method APES.
The computation complexity of different sampling models is given in Table \ref{table:time}. For a fair comparison, we use the same sampling size $M = 512$ and the same point embedding dimension of 128 in this table.

\subsection{Quantitative and Qualitative Results}
\begin{figure}[t]
    \centering
    \setlength{\abovecaptionskip}{0.2cm}
    \includegraphics[width=0.98\linewidth,trim=0 0 0 0,clip]{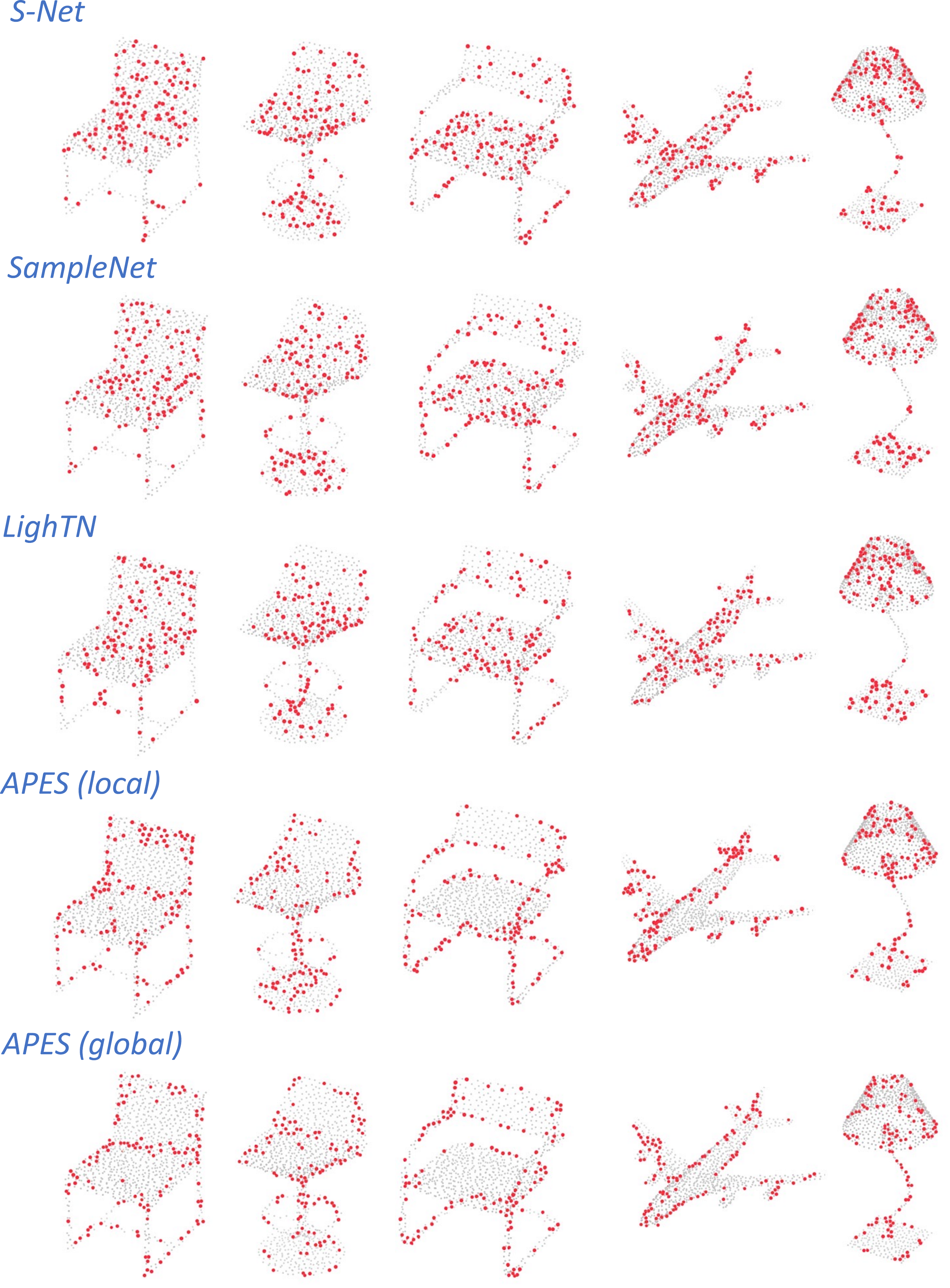}
    \caption{Qualitative comparison for the sampling of 128 points from input point clouds with 1024 points. 
    \vspace{-0.2cm}
    }
    \label{fig:vis_compare}
\end{figure}

Quantitative results are given in Table \ref{table:compare}, from which we can observe that both local-based and global-based APES achieve good classification results with the task network under different sampling ratios. Additional qualitative results are provided in Figure \ref{fig:vis_compare}. 
Although other learning-based methods achieve decent numerical results, it is difficult to identify their sampling patterns from the visualization results.
Their results look quite similar to random sampling. On the other hand, our proposed method shows a comprehensive sampling pattern of sampling point cloud outlines. 

\section{Conclusion}
\label{sec:conclusion}
In this paper, an attention-based point cloud edge sampling (APES) method is proposed. 
It uses the attention mechanism to compute correlation maps and sample edge points accordingly. Two variations of local-based APES and global-based APES are proposed based on different attention modes. Qualitative and quantitative results show that our method achieves excellent performance on common point cloud benchmark tasks.

For future work, it is possible to design other supplementary losses for the training. Moreover, we noticed that the different point distribution by edge point sampling hinders later upsampling operations and segmentation performance. It would be interesting to design upsampling methods that can better cope with edge point sampling.

{\small
\bibliographystyle{ieee_fullname}
\bibliography{main}
}

\end{document}